\documentclass[oneside,a4paper]{article}

\newif\ifdraft\draftfalse
\newif\ifinlineref\inlinereffalse
\newif\ifappendix\appendixtrue
\newif\ifdotikz\dotikzfalse
\dotikztrue

\usepackage{amssymb} %
\usepackage{amsthm}
\usepackage{amsmath}
\usepackage{url}
\usepackage{paralist}
\usepackage{booktabs}
\usepackage{multicol}
\usepackage{multirow}
\usepackage{array}
\usepackage[sort,numbers]{natbib}
\usepackage{fullpage}
\usepackage{graphicx}
\usepackage[normalem]{ulem}
\usepackage[utf8]{inputenc} %
\usepackage[T1]{fontenc} %

\ifdraft
\usepackage{todonotes}
\newcommand{\comment}[1]{\todo[inline,color=orange!40]{#1}}
\else
\newcommand{\comment}[1]{}
\fi

\ifdotikz
\usepackage{tikz}
\usetikzlibrary{shapes,arrows,backgrounds,%
matrix,patterns,arrows,decorations.pathmorphing,decorations.pathreplacing,%
positioning,fit,calc,decorations.text,shadows%
}
\pgfrealjobname{asp-teaching-hints-tr}
\else
\long\def\beginpgfgraphicnamed#1#2\endpgfgraphicnamed{\includegraphics{#1}}
\fi

\usepackage{mymacros}
\usepackage{myfigures}

\allowdisplaybreaks

\title{%
{\sc Technical Report:}\\
Giving Hints for Logic Programming Examples\\
without Revealing Solutions%
\thanks{This is a slightly extended English version of \cite{Avci2016siu} (original in Turkish).
This work has been supported by Scientific and Technological Research Council of Turkey (TUBITAK) Grant 114E777.}%
}

\author{Gokhan Avci, Mustafa Mehuljic, and Peter Schüller \\
Computer Engineering Department, Faculty of Engineering\\
Marmara University, Turkey \\
{\tt\small gokhan.avci@marun.edu.tr, mehuljic.mustafa@gmail.com, peter.schuller@marmara.edu.tr}}

\date{}

\begin{document}
\maketitle
\begin{abstract}
We introduce a framework for supporting learning
to program in the paradigm of Answer Set Programming (ASP),
which is a declarative logic programming formalism.
Based on the idea of teaching by asking the student
to complete small example ASP programs,
we introduce a three-stage method for
giving hints to the student without revealing the
correct solution of an example.
We categorize mistakes into
(i) syntactic mistakes,
(ii) unexpected but syntactically correct input,
and
(iii) semantic mistakes,
describe mathematical definitions of these mistakes,
and show how to compute hints from these definitions.
\end{abstract}

\ifdraft
\listoftodos
\fi

\section{Introduction}

Answer Set Programming (ASP)
\cite{Gelfond1988,Lifschitz2008,Gebser2012aspbook}
is a declarative logic programming
and knowledge representation paradigm.
Learning how to use ASP is challenging,
because logic programming is very different
from imperative programming (Java, C, C++)
and also different from Prolog
(in particular the order of rules
has no influence on the evaluation algorithm).

We here consider teaching ASP
by giving small real-world knowledge representation
examples and partial programs to students,
and ask them to complete the program.
To supporting the student,
examples can often be visualized and it is even possible
to connect visualization and knowledge representation
such that clicking the visualization can show
the parts of the program that define the visualized object \cite{Kloimullner2013}.
Creating such visualization is a work-intensive
task, and creating simple examples for teaching
ASP is also work-intensive.

To support the student in finding the correct answer,
in this work we describe a system for giving hints
based on the example program, the user input,
and a reference program,
such that the hints do not reveal the true answer.

Consider the following example specification.

\begin{quotation}{\it
We have a graph which represents the map of cities
connected by roads.  Cities are nodes, roads are edges,
all roads are bidirectional, and roads might be blocked.
The task is is to find a route from city $X$ to city $Y$
in the road network such that the path
uses only roads that are not blocked.
The graph is shown in Figure~\ref{figCities}
and is represented in the following set of facts.
}\end{quotation}
\begin{equation}
\label{eqExGiven}
  \begin{split}
  \begin{array}{@{\!\!}l@{\ }l@{}}
   \mathtt{road(istanbul,kocaeli).}
   &\mathtt{road(karabuk,bolu).}\\
   \mathtt{road(kocaeli,sakarya).}
   &\mathtt{road(duzce,karabuk).}\\
   \mathtt{blocked(duzce,zonguldak).}
   &\mathtt{road(bolu,zonguldak).}\\
   \mathtt{road(duzce,zonguldak).}
   &\mathtt{road(sakarya,duzce).}
  \end{array}
  \end{split}
\end{equation}
\begin{quotation}{\it
Moreover the fact that roads are bidirectional
is represented in the following rule.
}\end{quotation}
\begin{equation}
\label{eqSymmetry}
  \mathtt{road(X,Y)} \ \leftimpl \ \mathtt{road(Y,X).}
\end{equation}

\begin{quotation}
\noindent
\textbf{Question:}
{\it
Write a rule which defines predicate $\mathtt{open\_road(From, To)}$ that is true
for all pairs of cities where the direct road connection from $\mathtt{From}$ to $\mathtt{To}$ is not blocked.}
\end{quotation}

The expected answer to this example is as follows.
\begin{align}
  \label{eqExpected}
  \begin{array}{@{}r@{\ }l@{}}
  \mathtt{open\_road(X,Y)} \ \leftimpl \
    \mathtt{road(X,Y)},\ \nott \ \mathtt{blocked(X,Y)},\
  \nott \ \mathtt{blocked(Y,X).}
  \end{array}
\end{align}

However we might get several different kinds of answers
from students, for example the following.
\begin{align}
\label{syntax_err_1}
  &\mathtt{open\_road(X,Y) \ \leftimpl \ road(X,Y),\ \nott \ blocked(X,Y)} \\
\label{syntax_err_2}
  &\mathtt{open\_road(X,Y) \ \leftimpl \ road(X,Y),\ \nott \ blocked(X,Y);} \\
\label{preprocessing_err_1}
 &\mathtt{open\_road(X,Y)\ \leftimpl\ road(X,Y),\ \nott \ obstacle(X,Y).}  \\
\label{preprocessing_err_2}
 &\mathtt{open\_road(X,Y)\ \leftimpl\ road(X),\ \nott \ blocked(X,Y).} \\
\label{preprocessing_err_3}
 &\mathtt{open\_road(X,Y)\ \leftimpl\ road(x,y),\ \nott \ blocked(X,Y).} \\
\label{semantic_err_1}
 &\mathtt{open\_road(X,Y)\ \leftimpl\ road(X,Y),\ \nott\ blocked(duzce,bolu).}
\end{align}

Some of these answers are simply typos
or missing syntactical elements
such as \eqref{syntax_err_1} and \eqref{syntax_err_2}.
Other mistakes can be processed by ASP solvers
but do not yield the correct output
such as \eqref{preprocessing_err_1}--\eqref{semantic_err_1}.

\figureCitiesExample

In this work
we propose a framework that produces a hint
about the mistake from the student answer
and the reference program.
The hint does not reveal the correct solution.

This has the benefit,
that fewer examples are sufficient for teaching ASP
and fewer effort in visualization is required.
Moreover student motivation is not diminished by
giving away the solution.

We categorize mistakes into
syntactic mistakes such as \eqref{syntax_err_1}
and \eqref{syntax_err_2},
unexpected but syntactically correct input
such as \eqref{preprocessing_err_1}--\eqref{preprocessing_err_3},
and semantic mistakes
such as \eqref{semantic_err_1}.
We clearly define criteria for recognizing
each class of these mistakes,
and describe mathematically how to compute information
necessary to give a helpful hint,
including example hints for the case above.

We give preliminaries in Section~\ref{secPrelims},
describe the architecture of our hint giving framework
in Section~\ref{secFramework},
give details about mathematical classification of mistakes
and corresponding hints in Section~\ref{secLayers},
briefly mention related work in Section~\ref{secRelated}
and conclude in Section~\ref{secConclusion}.

\section{Preliminaries}
\label{secPrelims}

Answer Set Programming (ASP)
is a declarative logic programming paradigm
\cite{Gelfond1988,Lifschitz2008,Gebser2012aspbook}.
An atom is form $p(x_1,\ldots,x_l)$
with $0 \les l$,
and if $l\eqs0$ we write the atom short as $p$.
A program $P$ consists of a set of rules,
a rule $r$ is of the form
\begin{equation}
  \label{eqRule}
  \mathtt{
  \alpha_1 \lor \cdots \lor \alpha_k
  \leftimpl \beta_1, \ldots, \beta_n,
    \naf \beta_{n+1}, \ldots, \naf \beta_m
  }
\end{equation}
where $\alpha_i$ and $\beta_i$ are atoms,
called head and body atoms of $r$,
respectively.
We say that $H(r) = \{ \alpha_1,\ldots, \alpha_k \}$
is the \emph{head} of $r$, and
$B^+(r) = \{ \beta_1,\ldots,\beta_n\}$, respectively
$B^-(r) = \{ \beta_{n+1},\ldots,\beta_m\}$
are the positive, respectively
negative \emph{body} of $r$.
We call a rule a \emph{fact} if $m \eqs 0$,
\emph{disjunctive} if $k \gts 1$,
and a \emph{constraint} if $k \eqs 0$.
Atoms can contain constants, variables, and function terms,
and a program must allow for a finite instantiation.
Semantics of an ASP program $P$ is defined based
on the ground instantiation $\grnd(P)$
and Herbrand base $\HB_P$ of $P$:
an interpretation $I \ins \HB_P$ satisfies a rule $r$
iff $H(r) \caps I \neqs \emptyset$
or $B^+(r) \nsubseteqs I$
or $B^-(r) \caps I \neqs \emptyset$;
$I$ is a model of $P$ if it satisfies all rules in $P$.
The reduct $P^I$ of $P$ wrt.\ $I$
is the set of rules
$P^I \eqs \{ H(r) \leftimpl B^+(r) \mids B^-(r) \caps I \eqs \emptyset \}$
and an interpretation $I$ is an answer set
iff it is a $\subseteq$-minimal model of $P^I$.
Details of syntactic restrictions,
additional syntactic elements, and semantics
of ASP are described in the ASP-Core-2 standard~%
\cite{Calimeri2012}.

\figureFlowchart

\section{Hint Giving Framework}
\label{secFramework}

While writing programs students are making
different kind of mistakes.
Our methodology for giving hints
is shown in Figure~\ref{figFlowchart}.
The input consists of the student's answer program
$P_U$ and the reference solution $P_R$,
where we assume that all parts of the example
are combined into a single program.
Analyzing the input and giving hints
is based on three phases
which address different kinds of mistakes.

In the first phase, we consider syntactical errors.
If program is syntax-error free,
we check if the student used the vocabulary we expected.
In the third phase we evaluate the program,
compare it with an evaluation of the reference program,
and check if the output of both programs is the same.

In each step,
we potentially find a mistake,
produce a hint for the student,
and abort processing at that step.

We abort processing at the first phase where we find an error,
because each phase requires the previous one to complete.
For example, is impossible to perform analysis of rules
if the rules cannot be parsed,
i.e., are syntactically incorrect.

\section{Classification of Mistakes and Corresponding Hints}
\label{secLayers}

We next give details about each of the three phases.

\subsection{Syntactical Check}
\label{secSyntactic}

In this phase we are checking for syntax errors
in the input program.
A standard ASP parser will detect syntax errors
and according to the error we will give a hint.
For example the ASP grounder \gringo
gives the following error when parsing \ref{syntax_err_1}.
\begin{align*}
  \mathtt{{-}:3:8{-}9: syntax error, unexpected <EOF>}
\end{align*}

This indicates the location of the mistake in the source code
(line $3$, characters $8{-}9$)
and allows us to display the error
with additional visual support to the student,
for example as the following hint.
\begin{align*}
\begin{array}{@{}l@{}}
\mathbf{Hint:} \\
\mathtt{open\_road(X,Y)\ \leftimpl\ road(X,Y),\ not\ blocked(X,Y)}\\
\mathtt{\hspace*{22.00em}\uparrow}\\
\text{Syntax error, unexpected $<$EOF$>$.} \\
\text{Remember that rules are of the form} \\
\quad\mathtt{atom\ \leftimpl\ atom,\ not\ atom.} \\
\text{and atoms are of the form}\\
\quad\mathtt{predicate}\\
\text{or }\\
\quad\mathtt{predicate(arg1,arg2)}\\
\text{or similar.}
\end{array}
\end{align*}

If student's answer passes these two checks, third step is semantical check. In this phase we already have program which is free of syntax errors and passes preprocessing check. This time we will analyse user's answer set and compare it with answer set that is correct (e.g. key answer set) and give feedback according to this.

\subsection{Vocabulary Check}
\label{secVocabulary}

In this phase we parse the student's input program
and check if the expected predicates with expected arities
and the expected constants are used.
We formalize this mathematically.

Given a set of atoms
$A \eqs \{ p_1(t_1,\ldots,t_{k_1}), \ldots, p_m(t_1,\ldots,t_{k_m}) \}$
we define function
$\mi{preds}(A)$ which returns the set of predicates in $A$,
the function $\mi{predarities}(A)$
which returns a set of tuples of predicates with their arity
as occuring in $A$,
and the function $\mi{constants}(A)$
which returns the constants used in $A$.
\begin{align}
  && \mi{preds}(A) \eqs& \{ {p}&\mids&p(t_1,\ldots,t_k) \ins A \} && &\\
  && \quad\mi{predarities}(A) \eqs& \{ {(p,k)}\hspace*{-3ex}&\mids&p(t_1,\ldots,t_k) \ins A \} && &\\
  && \mi{constants}(A) \eqs& \{ {t_i}&\mids&p(t_1,\ldots,t_k) \ins A,\ 1 \les i \les k, \text{and } t_i \text{ is a constant term}\}. && \hspace*{2em} &
\end{align}

For example given the set of atoms
\begin{align*}
  A \eqs& \{ \mathtt{road(X,Y)}, \mathtt{road(X)}, \mathtt{blocked(x,y)} \} \\
\intertext{we obtain}
  \mi{preds}(A) \eqs& \{ \mathtt{road}, \mathtt{blocked} \}, \\
  \mi{predarities}(A) \eqs& \{ (\mathtt{road},1), (\mathtt{road},2), (\mathtt{blocked},2) \}, \text{ and} \\
  \mi{constants}(A) \eqs& \{ \mathtt{x}, \mathtt{y} \}.
\end{align*}

We next define functions for obtaining
information about predicates and their arities
in heads and bodies of rules and programs.
\begin{align*}
  \mi{bpreds}(r) \eqs& \mi{preds}(B^{+}(r) \cups B^{-}(r)) \\
  \mi{hpreds}(r) \eqs& \mi{preds}(H(r)) \\
  \mi{bpredarities}(r) \eqs& \mi{predarities}(B^{+}(r) \cups B^{-}(r)) \\
  \mi{hpredarities}(r) \eqs& \mi{predarities}(H(r)) \\
\intertext{where $r$ is a rule of form \eqref{eqRule}, moreover}
  \mi{bpreds}(P) \eqs& \bigcup_{r \ins P} \mi{bpreds}(r) \\
  \mi{hpreds}(P) \eqs& \bigcup_{r \ins P} \mi{hpreds}(r) \\
  \mi{bpredarities}(P) \eqs& \bigcup_{r \ins P} \mi{bpredarities}(r) \\
  \mi{hpredarities}(P) \eqs& \bigcup_{r \ins P} \mi{hpredarities}(r) \\
  \mi{preds}(P) \eqs& \mi{bpreds}(P) \cups \mi{hpreds}(P) \\
  \mi{predarities}(P) \eqs& \mi{bpredarities}(P) \cups \mi{hpredarities}(P)
\end{align*}
where $P$ is a program
--- a set of rules of form \eqref{eqRule}.

For example given rule \eqref{preprocessing_err_1}
we would have
\begin{align*}
  \mi{bpreds}(\eqref{preprocessing_err_1}) \eqs&
    \{ \mt{road}, \mt{blocked} \} \\
  \mi{hpreds}(\eqref{preprocessing_err_1}) \eqs&
    \{ \mt{open\_road} \} \\
  \mi{bpredarities}(\eqref{preprocessing_err_1}) \eqs&
    \{ (\mt{road},2), (\mt{blocked},2) \} \\
  \mi{hpredarities}(\eqref{preprocessing_err_1}) \eqs&
    \{ (\mt{open\_road},2) \}
\end{align*}

Given a reference solution $P_R$ and student input $P_U$
we can use the above definitions to give hints
without revewaling $P_R$.

We next use these definitions to compute
what is unexpected in a student's program
compared with the expected solution.

{\bf Usage of additional predicates.~~}
We can detect if the student
uses additional predicates
that are not in the reference implementation
by computing
\begin{align*}
  \mi{wrongpred}(P_U,P_R) = \mi{preds}(P_U) \setminus \mi{preds}(P_R).
\end{align*}

For example if
$P_R = \eqref{eqExGiven} \cup \eqref{eqSymmetry} \cup \eqref{eqExpected}$
and $P_U = \eqref{eqExGiven} \cup \eqref{eqSymmetry} \cup \eqref{preprocessing_err_1}$
then we obtain
$\mi{wrongpred}(P_U,P_R) = \{ \mt{obstacle} \}$. This allows us to easily produce the following hint.

\begin{quotation}
  \noindent
  \textbf{Hint:}
  Predicate $\mathtt{obstacle}$ should not be used.
\end{quotation}

Note that the hint does not reveal the sample solution.

{\bf Usage of predicates with wrong arity.~~}
We can detect if the student uses a predicate
with the wrong arity by computing
\begin{align*}
  \mi{wrongarity}(P_U,P_R) =
    \mi{predarities}(P_U) \setminus \mi{predarities}(P_R).
\end{align*}

For example with $P_U = \eqref{eqExGiven} \cup \eqref{eqSymmetry} \cup \eqref{preprocessing_err_2}$
we obtain
$\mi{wrongarity}(P_U,P_R) = \{ (\mt{road},1) \}$
and we can give the following hint accordingly.

\begin{quotation}
  \noindent
  \textbf{Hint:}
  Predicate $\mathtt{road}$ was used with arity $1$
  which is unexpected.
\end{quotation}

Note that if the student is unable to correct the problem
after this hint,
the next hint can reveal that the arity of $\mt{road}$
in the sample solution is $2$, without revealing the full
solution.

{\bf Usage of unexpected constants.~~}
We can detect if the student uses unexpected constants
by computing
\begin{align*}
  \mi{wrongcons}(P_U,P_R) =
    \mi{constants}(P_U) \setminus \mi{constants}(P_R).
\end{align*}

For example with $P_U = \eqref{eqExGiven} \cup \eqref{eqSymmetry} \cup \eqref{preprocessing_err_3}$
we obtain
$\mi{wrongcons}(P_U,P_R) = \{ \mt{x}, \mt{y} \}$
and we can give the following hint accordingly.

\begin{quotation}
  \noindent
  \textbf{Hint:}
  The program contains the following unexpected constants
  which are not required in the solution:
  $\mt{x}$, $\mt{y}$.
\end{quotation}

Note that this also shows if the student
has accidentally used constants instead of variables
as in \eqref{preprocessing_err_3}.

{\bf Further Hints.~~}
Using $\mi{bpreds}$, $\mi{hpreds}$, $\mi{bpredarities}$, and $\mi{hpredarities}$,
further hints that are more specific, can be produced,
for example that a certain predicate should be used
in the body of a rule in the solution.

All hints produced by this method are not revealing the solution,
they are just helping students to find out their mistakes
based on the vocabulary used in the program.
Students should find out what to do in order to fix this mistake.
When all sets are checked we can say that user's program is correct with respect to the expected vocabulary (which includes arities of predicates).

\subsection{Semantical Check}

The final phase is semantic verification:
we evaluate the student's program and the reference solution
using an ASP solver and verify if the results are the same.

If an answer set of student program includes
extra or missing entities,
the student program fails the check in this phase.

For example the answer \ref{semantic_err_1}
works correctly only for obstacles between Düzce and Bolu,
but not between Düzce and Zonguldak.
Given $P_R$ as above
and $P_U = \eqref{eqExGiven} \cup \eqref{eqSymmetry} \cup \eqref{semantic_err_1}$
we obtain one answer set $I_R \ins \AS(P_R)$
and one answer set $I_U \ins \AS(P_U)$
such that
$I_R \setminus I_U = \emptyset$
because the student's solution reproduces
all atoms that are true in the reference solution,
moreover
$I_U \setminus I_R = \{
  \mt{open\_road(duzce,zonguldak)},\allowbreak
  \mt{open\_road(zonguldak,duzce)} \}$
because the student's solution
additionally produces true atoms that should not be true.

The student program shows these blocked roads as open
which is the mistake.
By computing the difference between answer sets
we can give different levels of hints to the student,
for example the following generic hint.
\begin{quotation}
  \noindent
  \textbf{Hint:}
  The answer set contains more true atoms
  than it should.
\end{quotation}

If the student is not able to fix the problem,
we can also mention the predicate.
\begin{quotation}
  \noindent
  \textbf{Hint:}
  The answer set contains more true atoms
  of predicate $\mt{open\_road}$ than it should.
\end{quotation}

If even this does not help,
we can concretely say which true atom should not be true.
\begin{quotation}
  \noindent
  \textbf{Hint:}
  The answer set contains true atoms
  which should be false:
  $\mt{open\_road(duzce,zonguldak)}$
  and $\mt{open\_road(zonguldak,duzce)}$.
\end{quotation}

Again, these hints do not give solutions,
they guide the user to finding the problematic part of the
solution.

\section{Related Work}
\label{secRelated}

To the best of our knowledge there is no related work
focused on tools for teaching Answer Set Programming.
Programming Assignment Feedback System (PABS)
automatically generates feedback for programming assignments
in imperative languages based on impothe Java Virtual Machine,
however their feedback is based on plagiarism and failed unit tests
while our work gives detailed feedback within single unit tests
with additional constraint that we do not reveal the solution \cite{Ifflander2015}.

A systematic approach on teaching Prolog
based on a student model that takes into account experience from earlier courses
on imperative programming is presented in \cite{Stamatis2007}.
Misconceptions are separated in categories and different examples are given for each
category to facilitate in-depth understanding.
We use a similar approach to group mistakes and prepare hints,
however our work is on ASP which is more declarative than Prolog.
A didactic teaching and learning model for programming
is presented in \cite{Kaasboll1998}.
Our work can be seen as a teaching model for ASP.

Several approaches for \emph{debugging answer set programs}
exist, and these approaches could be integrated
into a teaching methodology.
Early work on this topic was limited to ground programs
and to explaining why an atom is present/absend
in an answer set \cite{Brain2005} using a procedural algorithm,
and later a more general explanation was achieved
in ground programs using declarative methods
similar to diagnosis \cite{Syrjanen2006}.
Pointing out reasons for a nonground ASP program
to not have a certain answer set
(unsatisfied rules, violated constraints,
unsupported atoms, and unfounded loops)
was developed in~\cite{Gebser2008},
refined in~\cite{Oetsch2010},
and later extended to choice rules, cardinality,
and weight constraints~\cite{Polleres2013}.
These debugging methods can be used
to realize a richer semantic check in our framework.

\section{Conclusion}
\label{secConclusion}

We have introduced a method for automatically giving hints
about Answer Set Programming exercises,
with the important property that the hints do not reveal
solutions, but guide the student in finding the solution.

Today methods from computer science are used in many different areas and therefore it
is an increasingly popular field of study. Consequently, more students are trained in basic
methods of computer science and programming. These students need to solve practical
programming assignments as part of their education. Feedback can help students to stay
motivated and assess their progress, but giving manual feedback is a very time consuming
task. Especially in large courses with beginners it is not practical due to staff limitations.
As a result it is desirable to automatically generate feedback whenever possible, so that
students can solve the easy problems with the help
of the computer and only require human assistance
for more difficult problems.

Regarding future work,
our syntactic check (Section~\ref{secSyntactic})
is based on existing ASP parsers,
and we augment the error message with helpful information.
An interesting future work
would be to create a parser that can suggest corrections
or produces more detailed analyses of the parser mistake.

\bibliographystyle{alpha}
\bibliography{logprog-teaching-hints}

\newcommand{\etalchar}[1]{$^{#1}$}
\begin{thebibliography}{KOPT13}

\bibitem[AMS16]{Avci2016siu}
Gokhan Avci, Mustafa Mehuljic, and Peter Sch\"uller.
\newblock {{\c{C}}{\"{o}}z{\"{u}}m{\"{u}} A{\c{c}}ığa {\c{C}}ıkarmadan
  Mantiksal Programlama {\"{O}}rneklerine İpucu Verme (Giving Hints For Logic
  Programming Examples Without Revealing Solutions)}.
\newblock In {\em Signal Processing and Communication Application Conference
  (SIU) (Sinyal İşleme ve İletişim Uygulamaları Kurultayı)}, pages
  513--516. IEEE, 2016.

\bibitem[BD05]{Brain2005}
Martin Brain and Marina {De Vos}.
\newblock {Debugging logic programs under the answer set semantics}.
\newblock In {\em International Workshop on Answer Set Programming}, volume
  142, pages 141--152. CEUR-WS.org, 2005.

\bibitem[CFG{\etalchar{+}}12]{Calimeri2012}
Francesco Calimeri, Wolfgang Faber, Martin Gebser, Giovambattista Ianni, Roland
  Kaminski, Thomas Krennwallner, Nicola Leone, Francesco Ricca, and Torsten
  Schaub.
\newblock {ASP-Core-2 Input language format}.
\newblock Technical report, ASP Standardization Working Group, 2012.

\bibitem[GKKS12]{Gebser2012aspbook}
Martin Gebser, Roland Kaminski, Benjamin Kaufmann, and Torsten Schaub.
\newblock {\em {Answer Set Solving in Practice}}.
\newblock Morgan Claypool, 2012.

\bibitem[GL88]{Gelfond1988}
Michael Gelfond and Vladimir Lifschitz.
\newblock {The Stable Model Semantics for Logic Programming}.
\newblock In {\em ICLP/SLP}, pages 1070--1080, 1988.

\bibitem[GPST08]{Gebser2008}
Martin Gebser, J{\"{o}}rg P{\"{u}}hrer, Torsten Schaub, and Hans Tompits.
\newblock {A meta-programming technique for debugging answer-set programs}.
\newblock In {\em AAAI Conference on Artificial Intelligence}, pages 448--453,
  2008.

\bibitem[IDBI15]{Ifflander2015}
Lukas Iffl{\"{a}}nder, Alexander Dallmann, Philip-Daniel Beck, and Marianus
  Ifland.
\newblock {PABS - a Programming Assignment Feedback System}.
\newblock In {\em Workshop "Automatische Bewertung von Programmieraufgaben"
  (ABP)}, pages 1--8, 2015.

\bibitem[Kaa98]{Kaasboll1998}
Jj~Kaasb{\o}ll.
\newblock {Exploring didactic models for programming}.
\newblock In {\em Norwegian Computer Science Conference (NIK)}, pages 195--203,
  1998.

\bibitem[KOPT13]{Kloimullner2013}
Christian Kloim{\"{u}}llner, Johannes Oetsch, Joerg Puehrer, and Hans Tompits.
\newblock {Kara: A System for Visualising and Visual Editing of Interpretations
  for Answer-Set Programs}.
\newblock In {\em INAP}, pages 325--344, 2013.

\bibitem[Lif08]{Lifschitz2008}
Vladimir Lifschitz.
\newblock What is {Answer Set Programming}?
\newblock In {\em AAAI Conference on Artificial Intelligence}, pages
  1594--1597, 2008.

\bibitem[OPT10]{Oetsch2010}
Johannes Oetsch, J{\"{o}}rg P{\"{u}}hrer, and Hans Tompits.
\newblock {Catching the Ouroboros: On Debugging Non-ground Answer-Set
  Programs}.
\newblock {\em Theory and Practice of Logic Programming}, 10(4-6):513--529,
  2010.

\bibitem[PFSF13]{Polleres2013}
Axel Polleres, Melanie Fr{\"{u}}hst{\"{u}}ck, Gottfried Schenner, and Gerhard
  Friedrich.
\newblock {Debugging non-ground ASP programs with choice rules, cardinality and
  weight constraints}.
\newblock In {\em International Conference on Logic Programming and
  Non-monotonic Reasoning (LPNMR)}, volume 8148 LNAI, pages 452--464, 2013.

\bibitem[SK07]{Stamatis2007}
Demosthenes Stamatis and Petros Kefalas.
\newblock {Logic Programming Didactics}.
\newblock In {\em Informatics Education Europe II Conference (IEEII)}, pages
  136--144, 2007.

\bibitem[Syr06]{Syrjanen2006}
Tommi Syrj{\"{a}}nen.
\newblock {Debugging Inconsistent Answer Set Programs}.
\newblock In {\em International Workshop on Non-Monotonic Reasoning (NMR)},
  pages 77----84, 2006.

\end{thebibliography}

\end{document}